\documentclass{article}

\usepackage[preprint,nonatbib]{neurips_2021}

\usepackage{graphicx}
\usepackage[bookmarks=false]{hyperref}
\usepackage{xspace}
\usepackage{subfigure}
\usepackage{enumerate}
\usepackage{enumitem}
\usepackage[utf8]{inputenc} % allow utf-8 input
\usepackage[T1]{fontenc}    % use 8-bit T1 fonts
\usepackage{hyperref}       % hyperlinks
\usepackage{url}            % simple URL typesetting
\usepackage{booktabs}       % professional-quality tables
\usepackage{amsfonts}       % blackboard math symbols
\usepackage{nicefrac}       % compact symbols for 1/2, etc.
\usepackage{microtype}      % microtypography
\usepackage{xcolor}         % colors
\usepackage{amsmath}

\makeatletter
\DeclareRobustCommand\onedot{\futurelet\@let@token\bmv@onedotaux}
\def\bmv@onedotaux{\ifx\@let@token.\else.\null\fi\xspace}
\def\eg{\emph{e.g}\onedot}

\def\etc{\emph{etc}\onedot} \def\vs{\emph{vs}\onedot}
\def\wrt{w.r.t\onedot}

\makeatother

\title{Graph Convolutional Network with Generalized Factorized Bilinear Aggregation} % for Text Classification}

\author{%
  Hao Zhu, Piotr Koniusz\thanks{The corresponding author. The code is available at \url{https://github.com/allenhaozhu/GFBP}} \\
  Australian National University and Data61/CSIRO\\
  Canberra, Australia \\
  \texttt{\{hao.zhu,piotr.koniusz\}@anu.edu.au} \\
}

\begin{document}

\maketitle

\begin{abstract}
Although Graph Convolutional Networks (GCNs) have demonstrated their power in various applications, the graph convolutional layers, as the most important component of GCN, are still using linear transformations and a simple pooling step. In this paper, we propose a novel generalization of Factorized Bilinear (FB) layer to model the feature interactions in GCNs. FB performs two matrix-vector multiplications, that is, the weight matrix is multiplied with the outer product of the vector of hidden features from both sides. However, the FB layer suffers from the quadratic number of coefficients, overfitting and the spurious correlations due to correlations between channels of hidden representations that violate the {\em i.i.d.} assumption. Thus, we propose a compact FB layer by defining a family of summarizing operators applied over the quadratic term. We analyze proposed pooling operators and motivate their use. 
Our experimental results on multiple datasets demonstrate that the GFB-GCN is competitive with other methods for text classification.
\end{abstract}

\section{Introduction}

%GCN is important
Text, as a weakly structured data,  is ubiquitous in e-mails, chats, on the web and in the social media, \etc. 
However, extracting information from text, is still a challenging task due to its weakly structured nature. 
Text classification, a fundamental problem in the Natural Language Processing (NLP), is used in document organization, news filtering, spam detection, and  sentiment analysis \cite{aggarwal2012survey}. 
With the rapid development of deep learning, text representations are often learnt by Convolutional Neural Networks (CNN) \cite{kim2014convolutional}, Recurrent Neural Networks (RNN) or Long Short-Term Memory (LSTM) networks.
%\cite{hochreiter1997long}. 
%to do: rewrite the next paragraph
As CNN and RNN are sensitive to locality and sequentiality,
%\cite {battaglia2018relational}, 
they capture well the semantic and syntactic information in local continuous word sequences. Conversely, CNN and RNN are worse at capturing the corpus with non-continuous long-time global word co-occurrence distance semantics \cite {peng2018large}.

Recently, Graph Convolutional Networks (GCN)  \cite{kipf2016semi,zhou2018graph,pmlr-v115-sun20a,kon_tpami2020a,zhu2021simple} have emerged as a powerful representation for the global structure of graphs and their node features. GCNs have been successfully used in the link prediction, graph classification and node classification. Approach \cite{yao2019graph}  successfully transformed the text classification into the graph node classification by employing a single text graph for a corpus based on word co-occurrence and document word relations. Furthermore, the vanilla GCN was used to generate the document representation of a node by aggregating word representation of its neighbours, and learning word embeddings at the same time.

GCNs learn to iteratively aggregate the hidden features of every node with its adjacent nodes in the graph to form new hidden features. There are two components in a graph convolutional layer: the transformation function and the aggregation function. The transformation function usually includes a learnable weight parameter for hidden features and a non-linearity. As an example, approach  \cite{xu2018powerful} employed a multi-layer perception as a transformation function. However, some works argue for lighter and simpler GCN variants, such as \cite{wu2019simplifying}, which drop the transformation function implicitly or explicitly.

An aggregation function often realizes a permutation-invariant function that captures so-called pooled statistics by sum-, mean-, max-pooling \etc. %, whose role is analogous to the role of local pooling in CNNs. 
%what's the problem
However, these operators are based on first-order statistics.
In contrast, bilinear pooling ~\cite{kon_tpami16} captures  second-order statistics, which represent better the underlying probability density function of data. 
%
% Second-order moments typically do not contain first-order statistics but their advantages compared to higher-order moments are a relative compactness.
%why we pay our attention to there
%
However, bilinear pooling applied to GCNs as a local pooling is costly due to its large number of parameters. % which may lead to overfitting.

In this paper, we introduce a Factorized Bilinear (FB) model that is generalized and compact, with the goal of  enhancing the capacity of graph convolutional layers in a simple manner. 
Specifically, as we compute FB, we multiply weight matrices from both sides with the outer product of the vector of hidden features, and we apply a row-wise  summarizaton function over such representations. 
Thus, instead of directly aggregating the document representation by word embeddings, we introduce second-order document representation resulting from adjacent hidden features of the previous layer. 
%rewrite
We note that the compact second-order representation resulting from the above step has the same size as its first-order counterpart, which constitutes an important computational improvement over a standard, quadratic in size, second-order representation. %Otherwise, learning a softmax classifier on a second-order representation has quadratic (rather than linear) cost \wrt the number of features. 
% However, second-order representation is usually very high dimensional, which can lead to overfitting or high computational cost for further layers like softmax.
%
%To address the above issue, we propose Factorized Bilinear model which performs a second-order factorization by multiplication of weight matrix on both sides of the graph matrix, and then it uses 
 a row-wise summarizaton function leading to a simple compact second-order representation. 

Finally, the compact second-order representation is linearly combined with its first-order counterpart, with the goal of capturing more statistics.  Our contributions are three-fold:
\renewcommand{\labelenumi}{\roman{enumi}.}
% \vspace{-0.4cm}
% \hspace{-1.0cm}
\begin{enumerate}[leftmargin=0.6cm]
    \item We introduce a Factorized Bilinear model into GCN that captures pairwise feature interactions and factorizes them by the multiplication of weight matrices on both sides with the outer product of the vector of hidden features. 
    \item For the FB model, to make it computationally applicable, we propose a novel family of summarization functions for that produce compact representations. We call this  Generalized  Factorized  Bilinear (GFB) model. We show that the sumamrization function in fact captures correlation of the feature \wrt all channels and can bee seen as low-rank modulator of the level of non-linearity of pooling. Our FB model can be easily incorporated into layers of GCN in contrast to traditional bilinear representations. 
    \item We validate the effectiveness of our approach on several standard benchmarks. Our proposed method achieves favourable performance compared to state-of-the-art methods with affordable complexity.
\end{enumerate}

\section{Related Work}
GCNs have drawn considerable attention due to their state of-the-art performance on graph-based tasks. Below, we summarize related works. %Studies on GNNs focus on extending the convolution and pooling operations known from CNNs to graphs.

\subsection{Text Classification based on Deep Learning}
Text classification approaches  can be divided into two categories: word embedding and deep learning approaches. 

Approaches across the first group build text representations on top of word embedding, which simplifies the downstream task. 
In the CBOW model of Word2Vec~\cite{mikolov2013distributed}, the model learns to predict a center word based on the context. 
PV-DM \cite{le2014distributed}, inspired by  Word2Vec, extends this idea to paragraphs by randomly sampling consecutive words from a paragraph, and predicting a center word from the randomly sampled set of words by taking as input the context words and a paragraph id.
Distributed Bag Of Words \cite{le2014distributed} (DBOW) %model is slightly different from the PVDM model.
 ignores the context words on the input but it forces the model to predict words that are randomly sampled from the paragraph on the output. 
However, these methods cannot learn word and document representation at the same time.

Another group of studies based on deep neural networks captures semantic and syntactic information in local consecutive word sequences. 
In~\cite{kim2014convolutional}, authors train a simple CNN with one layer of convolution on top of word vectors obtained from an unsupervised neural language model. 
Approach \cite{zhang2015character} extends this idea from words to characters.
However, the one layer CNN has a fixed receptive field, while Recurrent Neural Networks (RNNs) are able to learn temporal dependencies in the sequential data.
Researchers also introduced a generalization of the standard LSTM architecture to tree-structured network topology and shown its superiority over the sequential LSTM~\cite{tai2015improved} for representing the meaning of sentences.

\subsection{Graph Convolution Networks}
Convolution operation on graphs can be defined in either the spectral or non-spectral domain. 
Spectral approaches focus on redefining the convolution operation in the Fourier domain, utilizing spectral filters that use the graph Laplacian. 
Approach \cite{kipf2016semi} used a layer-wise propagation rule that simplifies the approximation of the graph Laplacian using the Chebyshev expansion \cite{defferrard2016convolutional}. 
The goal of non-spectral approaches is to define the convolution operation directly on graph nodes. 
Non-spectral approaches often assume that the central node aggregates features from adjacent nodes instead of defining the convolution operation in the Fourier domain. 
Generalizing the convolution operator to irregular domains is typically expressed as a neighborhood aggregation or message passing. 
There are two important components of GCNs: aggregation function and transformation. The former is usually a permutation invariant function \eg, sum, mean or maximum, whereas the later is linear or non-linear learner (\eg, Multi-Layer Perception). 
However, no GCN methods use pairwise feature interactions by considering second-order pooling.

Instead of learning node representation, approach \cite{zhang2020every} transforms the text classification into graph classification by forming graphs for different documents and learning the graph embedding for text classification. %Although this way can keep information for locality and sequentiality, it is difficult to handle moderate size dataset. 

%In this paper, we focus on node embedding based text classification.

\subsection{Higher-order Pooling}
Higher-order (beyond first order) statistics have been  studied in the context of texture recognition by the use of so-called Region Covariance Descriptors (RCD) \cite{tuzel_rc}. There exist several approaches for image retrieval and recognition, which perform some form of aggregation into second- or higher-order statistics extracted from local descriptors \cite{kon_tpami16} or the CNN convolutional feature maps \cite{lin2015bilinear,koniusz2018deeper}.  

However,  as second-order representations are usually vectorized, they result in representations of  $\mathcal{O}(d^2)$ size given $d$-dimensional feature vectors. 
Thus, second-order representations are substantially larger than the first-order representations.
In order to reduce the resulting memory cost, parameter explosion and over-fitting, we propose a simple and effective summarization method that produces representations of size equal to the size of first-order representations, while preserving the informativeness of second-order pooling.  
\cite{zhu2020bilinear,feng2021cross} give a simple method to reduce the size of representation by picking up the diagonal elements from high-order representations. In this paper, we propose a generalized framework and discuss four different methods to reduce the size of representations. 

\section{Preliminaries}
Below, we present the necessary prior techniques and notations, on which we build in this paper.

\subsection{Vanilla GCN for Text Classification}
\label{sec:NA}
Firstly, we introduce a generalized neighborhood aggregation for GCNs before introducing the Text GCN~\cite{yao2019graph} under such an aggregation scheme. 
Let $G = (V, E)$ be an undirected graph with node features $\mathbf{x}_v \in \mathbb{R}^{d_i}$ for $v \in V$, and $\tilde{G}$ be the graph obtained by adding a self-loop to every $v \in V$.
Here, $\mathbf{h}^{(l)}_v \in R^{d_h}$ represent hidden features of node $v$ learnt by the $l$-th layer of the model,
where $d_i$ and $d_h$ are  dimensions of the input and hidden features, respectively.
For the sake of clarity, we assume $d_h$ to be the same across layers and we simply refer to it as $d$. 
Let $\mathbf{h}^{(0)}_v = \mathbf{x}_v$ for the node features. 
%
% The neighborhood $N(v) = \{u \in V |(v, u) \in E\}$ of node $v$ is the set of adjacent nodes of $v$.
%
The neighborhood $\tilde{N}_{v} \equiv \{v\} \cup \{u \in V |(v, u) \in E\}$ on $\tilde{G}$ includes $v$.
For a model with $L$ layers, a typical neighborhood aggregation scheme  updates $\mathbf{h}^{(l)}_v$ for every $v\in V$ given $l$-th layer: %, $l\in\{ 1,\cdots,L\}$:
\vspace{-0.3cm}
\begin{equation}
    \mathbf{h}^{(l)}_v = \sigma\Big(\mathbf{W}_l \cdot \Omega \{\mathbf{h}^{(l-1)}_u, \forall u \in \tilde{N}_{v} \}\Big),
    \label{eq:neighbour_aggregation}
\end{equation}
where $\Omega$ is an aggregation function defined by the specific model, $\mathbf{W}_l$ is a trainable weight matrix on the $l$-th layer shared by all nodes, and $\sigma$ is a non-linear activation function. %, \eg a ReLU.

Graph Convolutional Networks (GCN) \cite{kipf2016semi} result from a specific realisation of $\Omega$, that is:
\begin{equation}
    \!\Omega(\{\mathbf{h}^{(l-1)}_u, \forall u \in \tilde{N}_{v} \})= \!\! \sum_{u \in \tilde{N}_{v}} (\text{deg}(v)\text{deg}(u))^{-\frac{1}{2}} \mathbf{h}^{(l-1)}_u,\!\!\!
    \label{eq:gcn_omega}
\end{equation}

%\vspace{-0.3cm}
\noindent{where} $\Omega$ sums  feature vectors of nodes $u\!\in \!\tilde{N}_v$ reweighted by the inverse square root of degrees of nodes $u$ and $v$. 
% Moreover, the full neighbour aggregation step for the vanilla GCN becomes:
% \begin{equation}
%         \mathbf{h}^{(l)}_v = \text{ReLU}\Big(\mathbf{W}_l \cdot\!\! \sum_{u \in \tilde{N}_{v}} (\text{deg}(v)\text{deg}(u))^{-\frac{1}{2}} \mathbf{h}^{(l-1)}_u\Big).
%         \label{eq:GCN}
% \end{equation}

% Text GCN can be defined as a simple two-layered vanilla GCN where the  node (word/document) embeddings of the second layer have the size equal the size of the label set. Thus, the first layer is defined as:
% \begin{equation}
%     \mathbf{h}_v^{(1)} = \text{ReLU}\Big(\sum_{u \in \tilde{N}_{v}} (\text{deg}(v)\text{deg}(u))^{-\frac{1}{2} } \text{row}_v(\mathbf{W}_1)\Big),
%     \label{eq:GCN_1st}
% \end{equation}
% where $\text{row}_v(\cdot)$ denotes the $v$-th row of the $\mathbf{W}_1$.
% Note that in the Text GCN, the input features are simply an identity matrix, thus the high-order aggregation in such a case is meaningless. The second layer is defined as:
% \begin{equation}
%         \mathbf{h}^{(2)}_v = \mathbf{W}_2 \cdot\!\! \sum_{u \in \tilde{N}_{v}} (\text{deg}(v)\text{deg}(u))^{-\frac{1}{2}} \mathbf{h}^{(1)}_u,
%         \label{eq:GCN_2nd}
% \end{equation}
% where $\sigma(x)=x$ as the output is combined with the softmax at this stage. %(\ie, %$\sigma(\mathbf{x}) = \text{softmax}(\mathbf{x})$).

\subsection{Bilinear Aggregation}
% Region Covariance Descriptors (RCDs), have proven effective to address visual recognition tasks. 
%
The bilinear (second-order) pooling~\cite{kon_tpami16,lin2015bilinear} replaces  fully connected layers by bilinear pooling that achieves remarkable improvements in the visual recognition by extracting second-order statistics from a set of feature vectors. %and adding a small positive value $\epsilon$ to the diagonal.
Specifically, if $\mathbf{h}_i\in \mathbb{R}^d$ corresponds to the $i$-th node embedding, node embeddings $\mathbf{H} = [\mathbf{h}_1,\cdots, \mathbf{h}_n ]\in \mathbb{R}^{d \times n}$  form an auto-correlation matrix of size $(d \times d)$ by:
\begin{equation}
    \mathbf{M} = \Big(\frac{1}{n}\sum_{i=1}^n\mathbf{h}_i\mathbf{h}_i^\top\Big) + \epsilon \mathbf{I},
\label{eq:cov}
\end{equation}
where $\mathbf{I}$ is the identity matrix and $\epsilon$ is a small reg. constant.

% But $A$ encodes second-order statistics, it completely discards the first-order ones, which may nonetheless bring valuable information. To keep the first order information, we add first order $\mu = \frac{1}{n}\sum_i^n x_i$ with A like the following form:
% \begin{equation}
% \[
% h=
% \left[
% \begin{array}{c c}
% A & \mu \\
% %\hline
% \mu^{\top} & 1
% \end{array}
% \right]
% \]
% \end{equation}

\section{Proposed Approach}
Below, we propose a second-order aggregation with the goal of replacing first-order pooling in GCNs. In Section \ref{sec:bp}, we outline Bilinear Pooling whose square number of coefficients is unacceptable due to high computational cost when multiplying with the same number of filter coefficient. In Section \ref{sec:fac}, we introduce  Factorized Bilinear Transformation which lets choose the number of filter coefficients by fetching the weight matrix inside the bilinear pooling step. In Section \ref{sec:vec}, we propose a Compact Vectorization by the matrix summarization function whose role is to further reduce the size of Factorized Bilinear representation to that of first-order approaches, and capture cross-channel correlations to mitigate the {\em i.i.d.} assumption on the feature distribution. In Section \ref{sec:ana}, our theoretical analysis shows that, in fact, our Compact Vectorization applied to Factorized Bilinear Transformation acts as a factorized attention, which can vary the level of non-linearity of the pooling operator.

\subsection{%Generalized Factorized
Bilinear Pooling}
\label{sec:bp}
Inspired by the second-order pooling, we replace the first-order aggregator in the generalized neighbour aggregation scheme by the second-order equivalent as follows:
\begin{equation}
    \mathbf{h}^{(l)}_v = \sigma\Big(\mathbf{W}_l \cdot \Omega\{\text{vec}(\mathbf{h}^{(l-1)}_u {\mathbf{h}^{(l-1)}_u}^\top\!+\epsilon \mathbf{I}), \forall u \in \tilde{N}_{v} \}\Big),
    \label{eq:bipool}
\end{equation}
where $\text{vec}(\cdot)$ extract the upper-diagonal from the matrix and stores it in a column vector. $\Omega$ is defined in Eq.~\ref{eq:gcn_omega}.

However, the use of auto-correlation matrix as a node representation is computationally inefficient due to the large size of $\mathbf{W}_l$ in Eq.~\ref{eq:bipool}, which leads to overfitting. Thus, we introduce a factorized model with a smaller number of parameters.

\subsection{Factorized Bilinear Transformation}
\label{sec:fac}
As the vectorization of auto-correlation matrix in Eq.~\ref{eq:bipool} results in $\mathcal{O}(d^2)$ coefficients, this indicates that for the representation $\mathbf{h}^{(l-1)}_u$ is of $16$ dimension size, the second-order representation is of $256$ dimensions ($136$ assuming coefficients of the  upper-triangular). Stacking few layers together results in the feature size  blown out of proportions. 

To tackle this issue, inspired by the factorization machine \cite{rendle2012factorization,li2017factorized}, Eq.~\ref{eq:bipool} can be reformulated as:
%\vspace{-0.3cm}
\begin{equation}
   \!\!\! \mathbf{h}^{(l)}_v\!=\!\sigma\Big(\Omega\{\text{vec}(\mathbf{W}_l \mathbf{h}^{(l-1)}_u  {\mathbf{h}^{(l-1)}_u}^\top\!\mathbf{W}_l^\top\!+\epsilon \mathbf{I}), \forall u\!\in\!\tilde{N}_{v} \}\Big),\!\!
    \label{eq:fm}
\end{equation}

\vspace{-0.0cm}
\noindent{where} $\mathbf{W}_l\!\in\!\mathbb{R}^{k\times d}$ and $\mathbf{h}^{(l-1)}_u\!\in\!\mathbb{R}^{d}$. We note ther reformulated Eq. \ref{eq:fm} uses  $\mathbf{W}_l$ matrix of size $k\times d$ rather than $d\times d$, which yields a vectorized representation of size $k^2$,  $k\!\ll\!d$. Although this  saves computations and reduces overfitting by low-rank approximation, compared to Eq.~\ref{eq:bipool}, it still requires matrix-vector multiplications which yield a representation of $k\times k$ such that $k^2\!\gg\!d$. %which is not directly applicable with standard loss functions such as Categorical Cross-Entropy loss, the softmax activation combined with the Cross-Entropy loss. The input should be logits representing unnormalized log probabilities. 
%Nonetheless, with some tweaking, factorized approaches can be combined with either a family of classifiers for factor machines or ordinary classifiers. %However, when the $\sigma$ function in Eq.~\ref{eq:fm} is softmax, the output is not meaningful in the probabilistic sense.
Thus, we consider below a compact summarization of matrices which further reduces the size of our representation.

\subsection{Generalized Compact Vectorization} 
\label{sec:vec}
%In \cite{yu2017second}, authors propose a operation of obtaining a vector $d$ with components from diagonal of $h(M)$ is done by
To reduce the size of our representation from $k^2$ to $k$, we propose a generalized compact vectorization of auto-correlation by defining a family of row-wise matrix summarization operators that capture statistics of auto-correlation matrix. We replace the vectorization operator in Eq.~\ref{eq:fm} by the following step:
%\vspace{-0.3cm}
\begin{equation}
% \textbf{GenVec}(\mathbf{M}) = \sum_{i=1}^k(\mathbf{e}_i^\top \mathbf{M} \mathbf{e'}_i) \mathbf{e}_i
\textbf{GenVec}(\mathbf{M}) = \sum_{i=1}^k g(\mathbf{e}_i^\top \mathbf{M}) \mathbf{e}_i,
\end{equation}

%\vspace{-0.1cm}
\noindent{where} $\mathbf{M}\in \mathbb{S}^{k}_{+}$ is an auto-correlation matrix, and  $\mathbf{e}_i\in \mathbb{R}^k$ is the $i$-th unit vector of the standard basis. 
%For $\mathbf{e'}_i$, we provide three different solutions which play different roles for $\mathbf{M}$.
For the function $g(\cdot)$, we provide four different candidates: (i) $g(\mathbf{x})\!=\!\max(\mathbf{x})$, (ii) $g(\mathbf{x})\!=\!\text{mean}(\mathbf{x})$, (iii) $g(\mathbf{x}; i)\!=\!x_i$,  and (iv) $g(\mathbf{x}; k)\!=\!\text{mean}(\text{Top}_k(\mathbf{x}))$, 
and we name them \textbf{MaxVec}, \textbf{MeanVec}, \textbf{DiagVec} and \textbf{TopkVec}, respectively. In the above definitions, $\mathbf(\cdot)$ and $\textbf{mean}(\cdot)$ are the maximum and mean over coefficients of a given vector, and $\textbf{Topk}$ returns top $k$ largest coefficients of a given vector.

\paragraph{Generalized Factorized Bilinear Graph Convolutional Network (GFB-GCN).}

Finally, our final GFB-GCN formulation combines the first- and second-order statistics as  follows:
%\vspace{-0.3cm}
\begin{equation}
    \mathbf{h}^{(l)}_v = \sigma\Big(\Omega\{\mathbf{W}^*_l \mathbf{h}^{(l-1)}_u + \lambda_l\textbf{GenVec}(\mathbf{M}), \forall u \in \tilde{N}_{v} \}\Big),
    \label{eq:cfm}
\end{equation}

%\vspace{-0.3cm}
\noindent{where}  $\mathbf{M}\!=\!\mathbf{W}_l \mathbf{h}^{(l-1)}_u  {\mathbf{h}^{(l-1)}_u}^\top\mathbf{W}_l^\top\!\!+\!\epsilon \mathbf{I}$ and we choose a desired summarization function $g(\cdot)$ according to four choices explained above. A learnable parameter $\lambda_l$ is the trade-off between the first- and second-order statistics.
%The operation can be presented also as
% \begin{equation}
% d=\sum_{i=1}^n(e_ie_i^\top)h(M)e_i    
% \end{equation}
% where matrix $P_i=e_ie_i^\top$ is the orthogonal projection matrix onto the line determined by standard basis vector $e_i$.

\paragraph{Implementation details.} 
%We define the a series of \textbf{GenVec} and observe although it can efficiently avoid the dimension explosion the computation of $\mathbf{M}$ is still computational costly. 
In practice, we avoid computing $\mathbf{M}$ explicitly due to high computational cost of outer product. Instead, if $\mathbf{h}'\!=\!\mathbf{W}_l \mathbf{h}^{(l-1)}_u$, then for MaxVec and MeanVec, we have:\\
 (i) $\textbf{MaxVec}(\frac{1}{n}\sum_u^n\mathbf{h'_u}\mathbf{h'_u}^\top)\!=\! \frac{1}{n}\sum_u^n\text{max}(\mathbf{h'_u})\mathbf{h'_u}$ and\\ (ii)  $\textbf{MeanVec}\left(\frac{1}{n}\sum_u^n\mathbf{h'_u}\mathbf{h'_u}^\top\right)\!=\!\frac{1}{n}\sum_u^n\text{mean}(\mathbf{h'_u})\mathbf{h'_u}$.\\
Computing \textbf{TopKVec} follows the same reasoning.
 % where $\text{mean}$ and $\text{max}$ are the functions to find the mean and max values among elements of the given vector $\mathbf{h_i}$.

\subsection{Analysis of Generalized Compact Vectorization with Factorized Bilinear Transformation}
\label{sec:ana}
Below, we demonstrate that our \textbf{MaxVec} pooling, our best performing pooling variant, performs two roles.

Firstly, \textbf{MaxVec} operator injects a non-linearity (piece-wise linear function to be precise) after the point-wise convolution of signal $\mathbf{h}^{(l-1)}_u$ with filters $\mathbf{W}_l\!=\!\left[\mathbf{w}_{1l},\cdots,\mathbf{w}_{kl}\right]^\top$  realized by the operation $\mathbf{W}_l \mathbf{h}^{(l-1)}_u$. The non-linearity following the convolution is required to attain the depth efficiency in neural networks, a property that lets neural nets realise increasingly more complex decision functions compared to decision functions of shallow networks. Notably, the level of non-linearity introduced by our summarizing operator can be varied by hidden features/filter weights (explained below).

Secondly, let $\mathbf{h}'\!=\!\mathbf{W}_l \mathbf{h}^{(l-1)}_u$.  Then, \textbf{MaxVec} operator can be seen as a confidence weighting which triggers the response proportional to probability of at least one correlation of $h'_i$ with $k$ convolutions of $\mathbf{h}^{(l-1)}_u$ with $\mathbf{w}_{1l},\cdots,\mathbf{w}_{kl}$ (stored as $\mathbf{h}'$) being observed. This confidence expresses our belief in whether a set of point-wise convolutions of the hidden feature vector with filters are important. The belief in a chosen feature increases
when it correlates with the maximum over all channels, as correlation `pools' a maximally correlated feature across channels to boost the response.

\textbf{MeanVec},  $\mathbf{W}_l \mathbf{h}^{(l-1)}_u \text{mean}\Big( {\mathbf{h}^{(l-1)}_u}^\top\!\mathbf{W}_l^\top\Big)\!=\!\mathbf{h}'\!\cdot\!\text{mean}( {\mathbf{h}'})$, simply captures the correlation of  point-wise convolutions $\mathbf{h}'\!=\!\mathbf{W}_l \mathbf{h}^{(l-1)}_u$ with their average, which introduces non-linearity, quantifies self-importance and, due to capturing the correlation with all channels, `pools' correlated features across channels, thus overcoming the {\em i.i.d.} assumption (also valid for MaxVec) on distribution of features across channels.

Let  probabilities of correlation of signal $\mathbf{h}^{(l-1)}_u$ (hidden feature vectors) with filters $\mathbf{w}_{1l},\cdots,\mathbf{w}_{kl}$ be expressed as  $p_i\propto h'_i$. We assume $h'_i\!=\!\mathbf{w}_{il}^\top\mathbf{h}^{(l-1)}_u\!\geq\!0$ to simplify our argumentation (negative values can be modeled too by splitting each variable into positive and negative halves, respectively). 
Then, we have the following set of inequalities:
\vspace{-0.2cm}
\begin{align}
& p_i\!\cdot\!\text{mean}\big(p_1,\cdots,p_k)\leq p_i\!\cdot\!\text{mean}\Big(\text{topk}\big(p_1,\cdots,p_k; k')\Big)\nonumber\\ &\leq p_i\!\cdot\!\text{max}\big(p_1,\cdots,p_k)\leq p_i\!\cdot\!\text{upper}\big(p_1,\cdots,p_k)\leq p_i,
\label{eq:ineq}
\end{align}

\vspace{-0.1cm}
\noindent{where} $\text{upper}\big(p_1,\cdots,p_k)=1\!-\!\prod_{j=1}^k(1\!-\!p_j)$ is {\em the probability of at least one $p$ from $p_1,\cdots,p_k$ being equal one}. We assume that $p_1,\cdots,p_k$ are drawn according to the i.i.d. principle from some underlying distribution. Of course, function $p_i\!\cdot\!\text{upper}\big(p_1,\cdots,p_k)$ is also non-linear in its nature. This implies that all our operators are non-linear as they are bounded (from bottom and top, respectively) by two concave operators, that is \textbf{MeanVec} and \textbf{UpperVec} realised by $\text{mean}(\cdot)$ and $\text{upper}(\cdot)$, respectively.

\textbf{MaxVec},  $\mathbf{W}_l \mathbf{h}^{(l-1)}_u \text{max}\Big( {\mathbf{h}^{(l-1)}_u}^\top\!\mathbf{W}_l^\top\Big)\!=\!\mathbf{h}'\!\cdot\!\text{max}( {\mathbf{h}'})$, has a tight upper bound $\text{upper}(\cdot)$,  
which implies that $\mathbf{h}'\!\cdot\!\text{max}( {\mathbf{h}'})$ captures simply the correlation of  point-wise convolutions $\mathbf{h}'\!=\!\mathbf{W}_l \mathbf{h}^{(l-1)}_u$ with the confidence of observing at least one stimulus $p_1,\cdots,p_k$ being equal one (being detected). Thus, if all $p_1,\cdots,p_k$ are fairly uniform and less then one, this formulation weights down the importance of entire \textbf{MaxVec} contribution due to lack of its distinctiveness.

\begin{figure*}[!htp]
\centering
\subfigure[MeanVec]{
\includegraphics[width=0.46\linewidth]{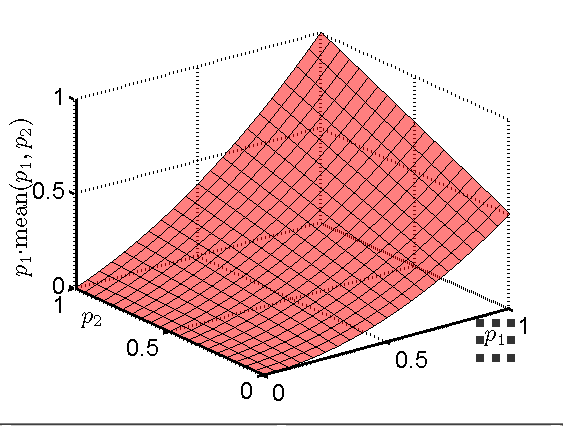}
\label{fig:mean}
}
\subfigure[MaxVec]{
\includegraphics[width=0.46\linewidth]{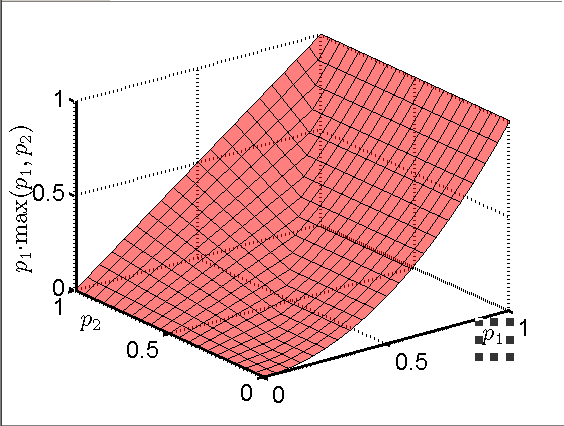}
\label{fig:max}
}
\subfigure[UpperVec]{
\includegraphics[width=0.46\linewidth]{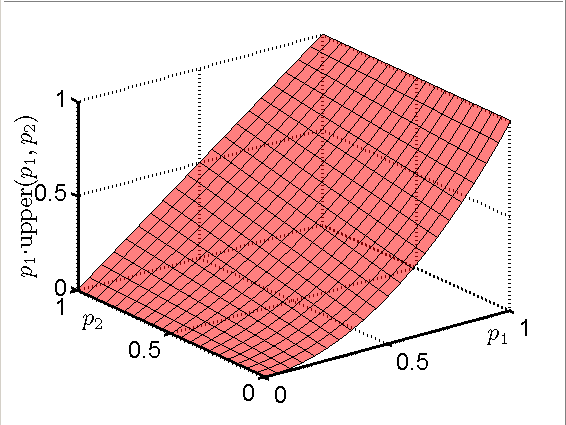}
\label{fig:upper}
}
\subfigure[Linear]{
\includegraphics[width=0.46\linewidth]{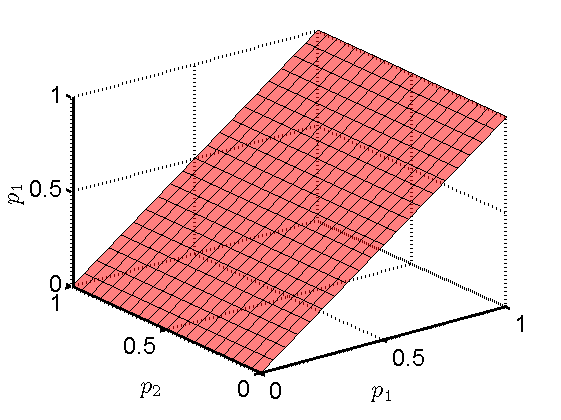}
\label{fig:linear}
}
\caption{Illustration of our pooling operators. Figures \ref{fig:mean}, \ref{fig:max}, \ref{fig:upper} and \ref{fig:linear} show MeanVec, MaxVec, UpperVec and the linear case, respectively.}
\label{fig:pools}
\end{figure*}
% \newcommand{\SrcImgHHH}{3.5cm}
% \begin{figure*}[t]%htbp % left bottom right top
% \centering%%%%
% %\vspace{-0.3cm}
% %
% \begin{subfigure}[0.245\linewidth]
% \centering\includegraphics[trim=5 5 5 5, clip=true, height=\SrcImgHHH]{mean.png}
% \caption{\label{fig:mean}}
% \end{subfigure}
% %
% \begin{subfigure}{0.245\linewidth}
% \centering\includegraphics[trim=5 5 5 5, clip=true, height=\SrcImgHHH]{max.png}
% \caption{\label{fig:max}}
% \end{subfigure}
% %
% %\\
% %
% \begin{subfigure}{0.245\linewidth}
% \centering\includegraphics[trim=5 5 5 5, clip=true, height=\SrcImgHHH]{upper.png}
% \caption{\label{fig:upper}}
% \end{subfigure}
% %
% \begin{subfigure}{0.245\linewidth}
% \centering\includegraphics[trim=5 5 5 5, clip=true, height=\SrcImgHHH]{linear.png}
% \caption{\label{fig:linear}}
% \end{subfigure}
% \vspace{-0.2cm}
% \caption{Illustration of our pooling operators. Figures \ref{fig:mean}, \ref{fig:max}, \ref{fig:upper} and \ref{fig:linear} show MeanVec, MaxVec, UpperVec and the linear case, respectively.}
% \label{fig:pools}
% \end{figure*}

Figure \ref{fig:pools} shows our summarizing operators. For the sake of clarity, we assume point-wise convolutions of some $\mathbf{h}^{(l-1)}_u$ with only two filters $\mathbf{w}_{1l}$ and $\mathbf{w}_{2l}$ which result in $p_1$ and $p_2$ proportional to correlation levels. %, as assumed earlier for simplicity. 
The role of each pooling operator is explained below.

Figure \ref{fig:mean} shows that \textbf{MeanVec} is a non-linear operator, however, as $p_2\!\rightarrow\!0$, even if $p_1\!\rightarrow\!1$, the final response $p_1\!\cdot\!\text{mean}\big(p_1,\cdots,p_2)$ drops to 0.5 which is an undesired effect as $p_1\!\rightarrow\!1$ indicates high confidence in filter $\mathbf{w}_{1l}$ being activated (correlated with the feature vector).

Figure \ref{fig:max} shows that \textbf{MaxVec} solves  the above issue \eg, if $p_1\!\rightarrow\!1$, then $p_1\!\cdot\!\text{max}\big(p_1,\cdots,p_2)\!\rightarrow\!1$, which reflects the strong confidence in the activation of filter $\mathbf{w}_{1l}$. %--this is the behavior we desire to propagate the resulting set of activations in the network. 
As $p_1\!<\!1$ and $p_2\!<\!1$, the activation of $\mathbf{w}_{1l}$  drops, as desired. Moreover, our confidence in activation depends on the probability of $p_1$ times the probability of at least one detection of among $p_1$ and $p_2$ being observed, which is proportional to $h'_1\!\cdot\!\text{max}( {\mathbf{h}'})$. This  mechanism  propagates the most informative activations to the next layer while suppressing uninformative activations.

\vspace{0.2cm}\hspace{-0.00cm}
\fbox{\begin{minipage}{39.0em}{
Notably, if $p_2\!=\!1$, \textbf{MaxVec} acts as a linear function \wrt $p_1$. This means the complexity of the signal is deemed low by GCN, due to the linear regime being selected.  For $p_2\!<\!1$, \textbf{MaxVec} becomes non-linear. The smaller the value of $p_2$, the greater the convexity, which validates our claim that the Factorized Bilinear model  acts as a low-rank modulator of level of non-linearity if paired with our \textbf{MaxVec} operator. %The level of non-linearity is varied according to $p_2$.
}
\end{minipage}
}
\vspace{0.1cm}

Figure \ref{fig:upper} shows that \textbf{UpperVec} indeed is an upper bound of \textbf{MaxVec}, which gives it a nice probabilistic interpretation  explained just below Eq. \ref{eq:ineq}. 

Figure \ref{fig:linear} shows the ordinary case, where the point-wise convolution is not followed by pooling. Then, $p_1$ simply behaves linearly (as it is not modulated by $p_2$), and $p_1$ is as an upper bound of \textbf{MaxVec} (and other operators). Alas, the linear setting violates the depth efficiency of neural nets.

%While \textbf{MaxVec} is non-linear, if $p_2\!\rightarrow\!1$, $p_1$ behaves linearly (the complexity of such signal is low thus linear behaviour is a desired property). However, if $p_2\!<\!1$, $p_1$ behaves non-linearly according to our confidence defined as 

%We also note that the probability of at least one correlation of $h^{(l-1)}_{iu}$ with $k$ filters $\mathbf{w}_{1l},\cdots,\mathbf{w}_{kl}$ being observed, which is realized approximately realized in \textbf{MaxVec}, ensures the propagation of the most informative activations to the next layer while suppressing uninformative activations. % (low confidence in hypothesis that point-wise convolutions $\mathbf{h}^{(l-1)}_{iu}$ with filters $\mathbf{w}_{1l},\cdots,\mathbf{w}_{kl}$ are discriminative).
%Do we need to extend the implementation and then analyze the complexity?
\subsection{Complexity Analysis}
% According to the definition of how weights $\mathbf{W}_l$ in Eq.~\ref{eq:cfm} interact with hidden features, the space complexity, which means the number of parameters for $\mathbf{W}_l$  in the GFB model, is $\mathcal{O}(kd)$, which is the same as for the vanilla GCN layer. The $\lambda_l$ is the only extra parameter, compared with the GCN layer. The computation cost of \textbf{MaxVec} is $\mathcal{O}(k+\log(k))$, \textbf{DiagVec} is $\mathcal{O}(k)$, \textbf{MeanVec} is $\mathcal{O}(2k)$, and \textbf{TopKVec} is $\mathcal{O}(k+k'\log(k))$. 
%Our summarization operates on $\mathbf{M} = \mathbf{W}_l \mathbf{h}^{(l-1)}_u  {\mathbf{h}^{(l-1)}_u}^\top\mathbf{W}_l^\top\!+\epsilon \mathbf{I}$. 
The computation cost of \textbf{MaxVec} is $\mathcal{O}(k)$ only as running maximum along $k$ dimensional vector $\mathbf{W}_l \mathbf{h}^{(l-1)}_u$ needs to touch $k$ elements. Moreover, \textbf{MeanVec} costs also $\mathcal{O}(k)$. However, the cost of \textbf{TopKVec} is $\mathcal{O}(k'+k\log(k))$ as the biggest cost is that incurred by the sorting function \eg, incremental sorting, but the standard practice is to provide the asymptotic cost which amounts to $\mathcal{O}(k'+k\log(k'))$. \textbf{DiagVec} cost is $\mathcal{O}(k)$.
The above costs represent the only extra computational cost incurred by us compared with the vanilla GCN layer. As a result, our model has close-to-linear/linear complexity \wrt $k$. We provide the runtime with the GFB layers in the experiments section.

\section{Experiments}
In this section, we evaluate the proposed method on three experimental tasks. We  (i) determine if our model achieves satisfactory results in text classification and brings advantages over Text GCN~\cite{yao2019graph} (vanilla GCN for text classification), and (ii) show that the proposed method is more efficient than other billinear pooling methods in text classification. We also provide evaluations on the entity recognition/graph classification in our \textbf{supplementary material}.
%\end{itemize}

\subsection{Setting}
\paragraph{Baselines.} We compare our method with  state of-the-art text classification and embedding methods such as: TF-IDF + LR, CNN~\cite{kim2014convolutional}, LSTM~\cite{liu2016recurrent}, PV-DBOW~\cite{le2014distributed}, PV-DM~\cite{le2014distributed}, PTE~\cite{tang2015pte}, fastText~\cite{joulin2016bag}, SWEM~\cite{shen2018baseline}, LEAM~\cite{wang2018joint}, Graph-CNN~\cite{defferrard2016convolutional}, Text GCN~\cite{yao2019graph}, SGC~\cite{wu2019simplifying} and GIN~\cite{xu2018powerful}. %Beside that, to evaluate the effectiveness and efficiency of the proposed method we also introduce

\vspace{0.05cm}
\noindent{\textbf{Architecture of GCN.}}
We found that a two-layer GCN performs better than a one layer GCN for the task of text classification, whereas adding more layer did not improve the performance. The paper on Text GCN and approach \cite{kipf2016semi} also note this. As input features  fed into the first layer are simply an identity matrix, the high-order aggregation in such a case is meaningless. Thus, in our two-layered architecture, the first layer follows Eq.~\ref{eq:neighbour_aggregation}, and the second layer uses variants such as GFB, Factorized Bilinear Pooling (FBP), Compact Billinear Pooling (CBP) and Bilinear Pooling (BP), described in the method section. 
\subsection{Datasets}
We ran our experiments on five widely used benchmark corpora including 20-Newsgroups (20NG), Ohsumed, R52 and R8 of Reuters 21578. We first preprocessed  datasets by cleaning and tokenizing text \cite{kim2014convolutional}. We then removed stop words defined in NLTK6 and low frequency words appearing less than 5 times for 20NG, R8, R52 and Ohsumed. 

% \paragraph{The 20NG dataset1}
% (bydate version) contains 18,846 documents evenly categorized into 20 different categories. In total, 11,314 documents are in the training set and 7,532 documents are in the test set.

% \paragraph{The Ohsumed corpus2} is from the MEDLINE database,
% which is a bibliographic database of important medical literature maintained by the National Library of Medicine. In this work, we used the 13,929 unique cardiovascular diseases abstracts in the first 20,000 abstracts of the year 1991. Each document in the set has one or more associated categories from the 23 disease categories. As we focus on single-label text classification, the documents belonging to multiple categories are excluded so that 7,400 documents belonging to only one category remain. 3,357 documents are in the training set and 4,043 documents are in the test set.

% \paragraph{R52 and R8}
% (all-terms version) are two subsets of the Reuters 21578 dataset. R8 has 8 categories, and was split to 5,485 training and 2,189 test documents. R52 has 52 categories, and was split to 6,532 training and 2,568 test documents.

% We first preprocessed all the datasets by cleaning and tokenizing text as~\cite{kim2014convolutional}. We then removed stop words defined in NLTK6 and low frequency words appearing less than 5 times for 20NG, R8, R52 and Ohsumed. 
%The statistics of the preprocessed datasets are summarized in Table~\ref{tab:datasets}.

\paragraph{Parameters.} We follow the setting of Text GCN~\cite{yao2019graph} that includes experiments on four widely used benchmark corpora such as 20-Newsgroups (20NG), Ohsumed, R52 and R8 of Reuters 21578.
For Text GCN and the proposed method, the embedding size of the first convolution layer is 200 and the window size is 20. We set the learning rate to 0.02, dropout rate to 0.5, the decay rate to 0, and $k=3$ for \textbf{TopkVec}. The 10\% of training set is randomly selected for validation. Following \cite{kipf2016semi}, we trained our method and Text GCN for a maximum of 200 epochs using the Adam~\cite{kingma2014adam} optimizer, and we stop training if the validation loss does not decrease for 10 consecutive epochs. The text graph was built according to steps detailed in \cite{yao2019graph}. %Sec.~\ref{sec:TG}. 

\begin{table*}[]
\centering
\caption{Test accuracy on the document classification task. Some Abbreviations: Factorized Bilinear Pooling (FBP), Compact Bilinear Pooling (CBP), Square Root Pooling (Square root), Weighted Average Pooling (WAP). For Bilinear Pooling (BP), we did not report results as experiments would require 20 days of 10 GPUs to run (\eg, 863.1 seconds per epoch, 200 epochs, 10 runs for 20NG). However, on R8 (smallest dataset), Text GCN+BP yields 0.9682$\pm$0.0042.}
\label{tab:result}
%\resizebox{\textwidth}{}{%
\begin{tabular}{|c|c|c|c|c|}
\hline
\textbf{Model}              & \textbf{20NG}            & \textbf{R8}              & \textbf{R52}             & \textbf{Ohsumed}                \\
\hline
TF-IDF + LR        & 0.8319 $\pm$ 0.0000 & 0.9374 $\pm$ 0.0000 & 0.8695 $\pm$ 0.0000 & 0.5466 $\pm$ 0.0000 \\
CNN-rand           & 0.7693 $\pm$ 0.0061 & 0.9402 $\pm$ 0.0057 & 0.8537 $\pm$ 0.0047 & 0.4387 $\pm$ 0.0100  \\
CNN-non-static     & 0.8215 $\pm$ 0.0052 & 0.9571 $\pm$ 0.0052 & 0.8759 $\pm$ 0.0048 & 0.5844 $\pm$ 0.0106  \\
LSTM               & 0.6571 $\pm$ 0.0152 & 0.9368 $\pm$ 0.0082 & 0.8554 $\pm$ 0.0113 & 0.4113 $\pm$ 0.0117  \\
PV-DBOW            & 0.7436 $\pm$ 0.0018 & 0.8587 $\pm$ 0.0010 & 0.7829 $\pm$ 0.0011 & 0.4665 $\pm$ 0.0019  \\
PV-DM              & 0.5114 $\pm$ 0.0022 & 0.5207 $\pm$ 0.0004 & 0.4492 $\pm$ 0.0005 & 0.2950 $\pm$ 0.0007  \\
PTE                & 0.7674 $\pm$ 0.0029 & 0.9669 $\pm$ 0.0013 & 0.9071 $\pm$ 0.0014 & 0.5358 $\pm$ 0.0029 \\
fastText           & 0.7938 $\pm$ 0.0030 & 0.9613 $\pm$ 0.0021 & 0.9281 $\pm$ 0.0009 & 0.5770 $\pm$ 0.0049 \\
SWEM               & 0.8516 $\pm$ 0.0029 & 0.9532 $\pm$ 0.0026 & 0.9294 $\pm$ 0.0024 & 0.6312 $\pm$ 0.0055  \\
LEAM               & 0.8191 $\pm$ 0.0024 & 0.9331 $\pm$ 0.0024 & 0.9184 $\pm$ 0.0023 & 0.5858 $\pm$ 0.0079  \\
Graph-CNN        & 0.8142 $\pm$ 0.0032 & 0.9699 $\pm$ 0.0012 & 0.9275 $\pm$ 0.0022 & 0.6386 $\pm$ 0.0053 \\
Text GCN (WAP)           & 0.8597 $\pm$ 0.0014 & 0.9707 $\pm$ 0.0010 & 0.9356 $\pm$ 0.0018 & 0.6836 $\pm$ 0.0056  \\
\hline
Text SGC (WAP)  & 0.8853 $\pm$ 0.0005 & 0.9723 $\pm$ 0.0002 & 0.9403 $\pm$ 0.0003 & 0.6853 $\pm$ 0.0003 \\
Text GIN (WAP) &0.8414 $\pm$ 0.0084 &0.9352 $\pm$ 0.0154 &0.9314 $\pm$ 0.0074 &0.6323 $\pm$ 0.0098\\
\hline
FBP (Eq.\ref{eq:fm}) & 0.8426 $\pm$ 0.0021 &0.9659 $\pm$ 0.0045  &0.9370 $\pm$ 0.0097, & 0.6601 $\pm$ 0.0050 \\
CBP & 0.7789 $\pm$ 0.0078 &0.9404 $\pm$ 0.0203  &0.8638 $\pm$ 0.0900, &0.5986 $\pm$ 0.0098 \\
Text GCN + Square root & 0.8618 $\pm$ 0.0013  &0.9700 $\pm$ 0.0011 &0.9383 $\pm$ 0.0013 &0.6915 $\pm$ 0.0029 \\
\hline
GFBP+MeanVec            & 0.8657 $\pm$ 0.0018 & 0.9744 $\pm$ 0.0017 & 0.9424 $\pm$ 0.0009 & 0.6943 $\pm$ 0.0039  \\
GFBP+DiagVec            & 0.8682 $\pm$ 0.0013 & 0.9767 $\pm$ 0.0011 & 0.9450 $\pm$ 0.0011 & 0.7164 $\pm$ 0.0023  \\
GFBP+MaxVec            & \textbf{0.8718 $\pm$ 0.0017} & \textbf{0.9770 $\pm$ 0.0009} & 0.9448 $\pm$ 0.0016 & \textbf{0.7174 $\pm$ 0.0022}  \\
GFBP + TopKVec & 0.8692 $\pm$ 0.0015 &0.9756 $\pm$ 0.0024  &\textbf{0.9466 $\pm$ 0.0019}, &0.7107 $\pm$ 0.0021 \\
\hline
\end{tabular}
%}
\end{table*}

% \textbf{20NG}  & 0.8426 $\pm$ 0.0021 &   \textbf{0.8682 $\pm$ 0.0013} \\
% \textbf{R8}  & 0.9659 $\pm$ 0.0045 &   \textbf{0.9767 $\pm$ 0.0011} \\
% \textbf{R52}  & 0.9370 $\pm$ 0.0097 &   \textbf{0.9450 $\pm$ 0.0011} \\
% \textbf{Ohsumed}   & 0.6601 $\pm$ 0.0050 &   \textbf{0.7164 $\pm$ 0.0023} \\
\subsection{Results}
Table~\ref{tab:result} presents the test accuracy of each model. Our method significantly outperforms all other models which showcases the effectiveness of our method on long text datasets. 
We notice that for Ohsumed and R52, the improvements are relatively greater than on other datasets and thus we analyze these results in our supplementary material. We also provide other results based on variants of Text GCN with SGC~\cite{wu2019simplifying} and GIN~\cite{xu2018powerful}. Note that SGC do not have hidden layer, thus the embedding of words and documents for 20NG are over 10000 size, which makes SGC gain advantage.

To showcase the effectiveness of our idea, we also provide comparisons to bilinear aggregation (\eg, Eq.~\ref{eq:bipool}), compact billinear pooling, and factorized bilinear pooling \cite{li2017factorized}. Because bilinear aggregation is extremely slow (about one month to finish evaluations on four datasets), we only provide a single result on R8, that is 0.9682$\pm$0.0042.

\paragraph{Speed Analysis}
We show the runtime speed comparison of a two-layer GCN %with and without FB layers 
in Table~\ref{tab:speed}. The test is performed on the GTX1080 GPU. From Table~\ref{tab:speed}, we can  see that all factorized methods (including FBP) are more efficient than bilinear pooling, which is two orders of magnitude slower compared to first-order methods. Compared with the vanilla text GCN, the proposed methods only slightly increase computation cost (20\%-30\%). The GFBP+\textbf{DiagVec} is the fastest while GFBP+\textbf{MaxVec} is marginally slower but provides overall best results. We also provide the speed of Compact Bilinear Pooling (CBP). Although CBP is just slightly slower than our method, its recognition performance is not competitive.

\begin{figure}[h]
\centering
\includegraphics[width=0.80\textwidth]{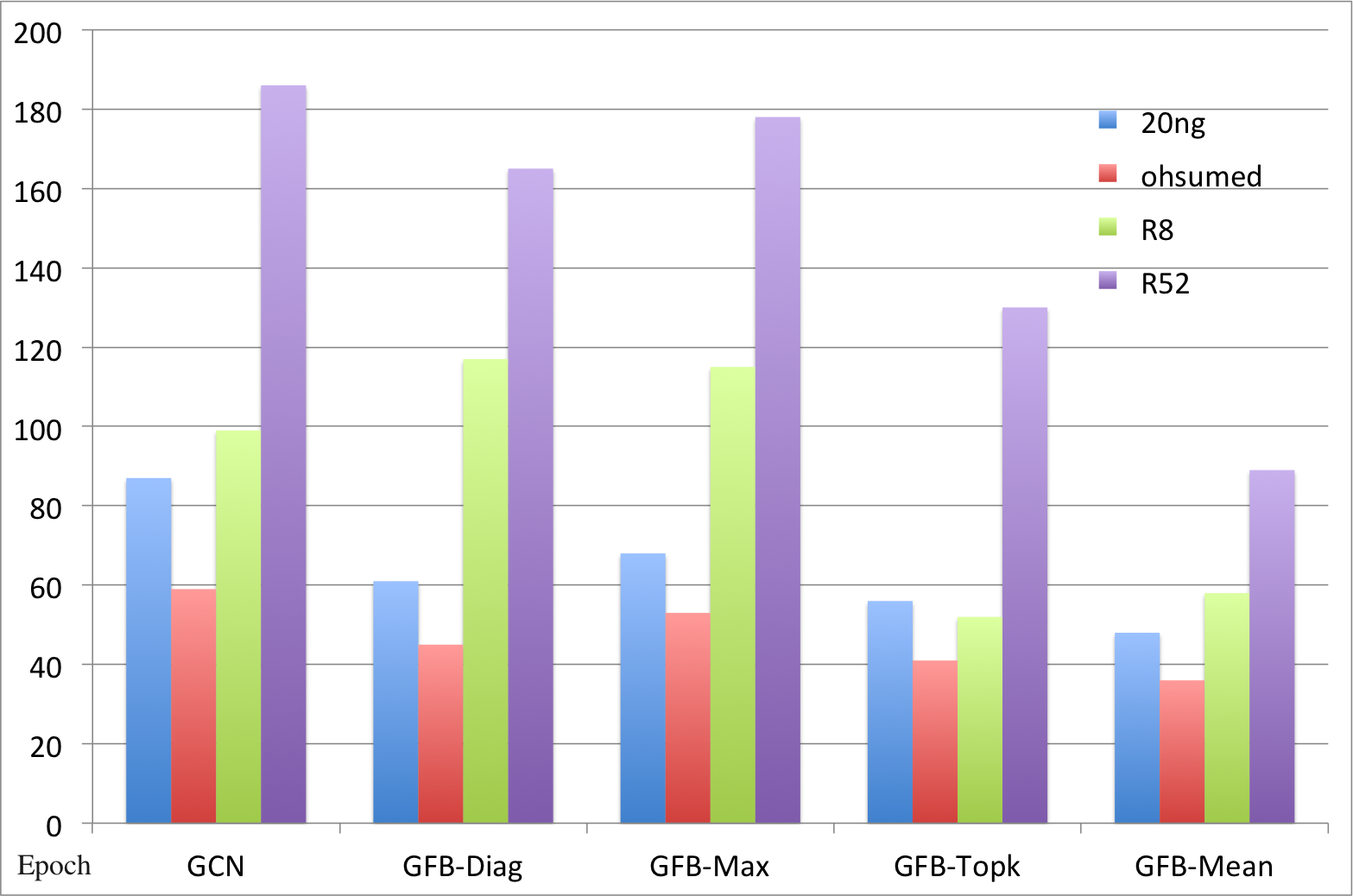}
\caption{The number of epochs at which the early stopping criterion was met (proposed methods \vs the vanilla GCN).}
\label{fig:convergence}
\end{figure}

% For Inception-BN, the loss of speed is still tolerable. In addition, since we only insert a single FB block, it has little impact on the speed of large networks, e.g. ResNet-1001. Lastly, cuDNN accelerates all
% methods a lot. We believe that the training speed of our FB layers will also benefit from a deliberate optimization.
\begin{table}[]
\centering
\caption{Time cost per epoch during training on the document classification task (in seconds).}
\label{tab:speed}
% \resizebox{\textwidth}{!}{%
\begin{tabular}{|c|c|c|c|c|}
\hline
\textbf{Model}              & \textbf{20NG}            & \textbf{R8}              & \textbf{R52}             & \textbf{Ohsumed}                \\
\hline
Text GCN           & 13.90  & 0.998  &  1.673  & 3.300   \\
FBP (Eq.\ref{eq:fm}) &15.63 &1.242 &2.317 & 3.907 \\
CBP &23.09&1.664&2.531& 4.913\\
BP (Eq.\ref{eq:bipool}) & 863.1 & 111.6 & 126.4 & 261.8 \\
\hline
GFBP+MeanVec            & 16.80  & 1.207 & 2.057 & 3.999  \\
GFBP+DiagVec            & 14.98 & 1.143 & 2.097 & 3.608 \\
GFBP+MaxVec            & 18.39 & 1.248 & 2.076 & 3.933  \\
GFBP+TopKVec           & 23.34 & 2.032 & 8.707 &  6.607 \\
\hline
\end{tabular}
% }
\end{table}
\paragraph{Fast Convergence of GCN with Generalized Factorized Model.} 
Below, we compare the convergence of our Generalized Factorized Model against the Vanilla GCN. Fig.~\ref{fig:convergence} demonstrates the number of epochs for each method till early stopping criterion is met (no further increase in the classification performance during validation). The GFBP+\textbf{MeanVec} and \textbf{TopKVec} converge much faster than other methods. The other two variants of GFBP are also competitive compared to the vanilla GCN, although for R52, they need few more epochs than the Text GCN. On other datasets, these variants still converge faster than the TextGCN. 
% More specifically, we also present the convergence curves of GFBP+\textbf{MeanVec} for the text classification in Fig.\ref{fig:comp}. 
Compared with the Text GCN, the convergence of the proposed method is significantly faster as our method converges in less than 30 epochs (32 epochs  on ohsumed dataset). At the same time, Text GCN runs until the 68 epoch (the early stopping condition). This trend is consistent on all four datasets. The faster convergence stems from the ability of our sumamrization operators to `pool' features between channels (thus reducing degrees of freedom that stem from the correlation of features in different channels) \cite{koniusz2018deeper}. 

\section{Discussion and Conclusions}
In this paper, we have proposed a novel neighbourhood aggregation scheme for GCNs and showcased results on the Text GCN.  Instead of using first-order local average pooling, we have introduced a generalized factorized bilinear neighbourhood aggregation scheme with summarization operators that attain compactness on the par of first-order average pooling, exploit correlation between channels, and realize low-rank modulation of pooling to interpolate between linear activation (simple signals) and various levels of non-linear activations realized by our operator. Our \textbf{supplementary material} presents further results on text classification (different metrics),  entity  recognition  and graph classification.

\appendix
\section{Text Classification}
\subsection{Text Graph}
\label{sec:TG}
To convert text classification into the node classification on graph, there are two relationships
considered when forming graphs: (i) the relation between documents and words and (ii) the connection between words.
For the first type of relations, we build edges among word nodes and document nodes based on the word occurrence in documents. 
The weight of the edge between a document node and a word node is the Term Frequency-Inverse Document Frequency \cite{rajaraman2011mining} (TF-IDF) of the word in the document applied to build the Docs-words graph. 
For the second type of relations, we build edges in graph among word co-occurrences across the whole corpus. 
To utilize the global word co-occurrence information, we use a fixed-size sliding window on all documents in the corpus to gather co-occurrence statistics.
Point-wise Mutual Information \cite{church1990word} (PMI), a popular measure for word associations, is used to calculate weights between two word nodes according to the following definition:
\begin{equation}
\text{PMI}(i,j) = \log\frac{p(i,j)}{p(i)p(j)}
\end{equation}
where $p(i,j)=\frac{W(i,j)}{W}$, $p(i)=\frac{W(i)}{W}$. $W(i)$ is the number of sliding windows in a corpus that contain word $i$, $W(i, j)$ is the number of sliding windows that contain both word $i$ and word $j$, and $W$ is the total number of sliding windows in the corpus.
A positive PMI value implies a high semantic correlation of words in a corpus, while a negative PMI value indicates little or no semantic correlation in the corpus. 
Therefore, we only add edges between word pairs with positive PMI values:
\[
\mathbf{A}=
\left[
\begin{array}{c|c}
\mathbf{W}_1 & \mathbf{W}_2 \\
\hline
\mathbf{W}_2^{\top} & \mathbf{I}
\end{array}
\right]
\]

or

\begin{equation}
A_{ij}=\left\{
\begin{aligned}
 \text{PMI}(i,j) & \quad\text{if } i,j~\text{are words, PMI}(i,j) > 0, \\
 \text{TF-IDF}_{ij} & \quad\text{if } i~\text{is document},j~\text{is word}, \\
  1 & \quad\text{if } i=j, \\
  0 & \quad\text{otherwise}.
\end{aligned}
\right.
\end{equation}

\subsection{Relationship to Factorization Machines}
Factorization Machine (FM) \cite{rendle2012factorization} is a popular predictor in machine learning and data mining, especially for very sparse data. Factorized Bilinear (FB) Model ~\cite{li2017factorized} is much more general, which can be integrated into regular neural networks seamlessly for different kinds of tasks because it extends FM to multi-dimensions.  %rather than 2-way. 
In this way, a 2-way FM is a special case of our FB model. Our approach, Generalized Factorized Billinear Model, is more general and compact than the FM model \eg, we can modulate the level of non-linearity in the response. As an example, we can rewrite the transformation part with \textbf{DiagVec} as: % a similar FB form:
\begin{equation}
    y = \mathbf{W} \mathbf{x} + \mathbf{F}\mathbf{x}\otimes  (\mathbf{F}\mathbf{x})^\top ,\; \text{s.t. } \mathbf{F} = \lambda\mathbf{W},
\end{equation}
where we suppress learning a new parameter matrix $\mathbf{F}$ for feature interactions. Instead of it, we only add a differentiable parameter $\lambda$ and a constraint as $\mathbf{F} = \lambda\mathbf{W}$. In the experiments, we  demonstrate that our proposed method is more efficient and effective.

\begin{table*}[]
\caption{The statistics of datasets.}
\centering
\begin{tabular}{c|c|c|c|c|c|c|c}
\hline
\textbf{Dataset} & \textbf{\# Docs} & \textbf{\# Training} & \textbf{\# Test} & \textbf{\# Words} & \textbf{\# Nodes} & \textbf{\# Classes} & \textbf{Average Length} \\
\hline
20NG    & 18,846  & 11,314      & 7,532   & 42,757   & 61,603   & 20         & 221.26         \\
R8      & 7,674   & 5,485       & 2,189   & 7,688    & 15,362   & 8          & 65.72          \\
R52     & 9,100   & 6,532       & 2,568   & 8,892    & 17,992   & 52         & 69.82          \\
Ohsumed & 7,400   & 3,357       & 4,043   & 14,157   & 21,557   & 23         & 135.82         \\
\hline
\end{tabular}
\label{tab:datasets}
\end{table*}
\subsection{Datasets}
We ran our experiments on five widely used benchmark corpora including 20-Newsgroups (20NG), Ohsumed, R52 and R8 of Reuters 21578.

\paragraph{The 20NG dataset1}
(bydate version) contains 18,846 documents evenly categorized into 20 different categories. In total, 11,314 documents are in the training set and 7,532 documents are in the test set.

\vspace{0.1cm}
\noindent{\textbf{The Ohsumed corpus2}}, a part of the MEDLINE database,
 is a bibliographic database of important medical literature maintained by the National Library of Medicine. In this work, we used the 13,929 unique cardiovascular diseases abstracts in the first 20,000 abstracts of the year 1991. Each document in the set has one or more associated categories from the 23 disease categories. As we focus on single-label text classification, the documents belonging to multiple categories are excluded so that 7,400 documents belonging to only one category remain. 3,357 documents are in the training set and 4,043 documents are in the test set.

\paragraph{R52 and R8}
(all-terms version) are two subsets of the Reuters 21578 dataset. R8 has 8 categories, and was split to 5,485 training and 2,189 test documents. R52 has 52 categories, and was split to 6,532 training and 2,568 test documents.

\subsection{Data Imbalance}
Our method outperforms the Text GCN in terms of accuracy. However, the definition of accuracy for multi-class tasks (a simple average) may disadvantage algorithms in presence of imbalanced domains. R52 is the most imbalanced datasets for which the number of samples in each category varies from 1 to 1083. Thus, we analyze results in terms of Macro-Precision, Macro-Recall and Macro-F1. The results in Table~\ref{tab:R52} show significant improvements in terms of macro-recall and macro-F1, which shows that our method can classify correctly samples belonging to categories with very limited training samples. We also obtain the similar result on Ohsumed (in Table.~\ref{tab:ohsumed}), where the number of samples in each category varies from 10 to 600, although the number of categories is  below the number of categories of  R52. 
\begin{table}[]
\centering
\caption{Macro-Precision, Recall, F-1 Results on R52.}
\label{tab:R52}
\begin{tabular}{|l|l|l|l|}
\hline
R52      & Precision & Recall & F1     \\
\hline
Text GCN & 0.7561    & 0.6694 & 0.6864 \\
\hline
%Ours     & 0.7437    & 0.6935 & \textbf{0.7002} \\
GFB+MeanVec     & 0.7654   & \textbf{0.6864} & \textbf{0.7058} \\
GFB+DiagVec     & 0.7662   & 0.6747 & 0.6925 \\
GFB+MaxVec     & 0.7598   & 0.6845 & 0.6980 \\
GFB+TopKVec &\textbf{0.7741}& 0.6834& 0.7008\\
\hline
\end{tabular}
\end{table}

\begin{table}[]
\centering
\caption{Macro-Precision, Recall, F-1 Results on Ohsumed.}
\label{tab:ohsumed}
\begin{tabular}{|l|l|l|l|}
\hline
Ohsumed      & Precision & Recall & F1     \\
\hline
Text GCN & 0.6814    & 0.5913 & 0.6216 \\
\hline
%Ours     & 0.7437    & 0.6935 & \textbf{0.7002} \\
GFB+MeanVec     & 0.6606   & 0.6110 & 0.6255 \\
GFB+DiagVec     & 0.6873   & \textbf{0.6336} & \textbf{0.6517} \\
GFB+MaxVec     & \textbf{0.6938}   & 0.6279 & 0.6510 \\
GFB+TopKVec  &0.6813& 0.6256& 0.6446 \\
\hline
\end{tabular}
\end{table}

\section{Entity Classification}
Below, we study the task of entity classification in a knowledge base. In order to infer, for example, the type of an entity (\eg, person or company), a successful model needs to reason about the relations with other entities that the main entity is involved in.

\subsection{Extension of R-GCN to Generalized Factorized Bilinear R-GCN (GFB-R-GCN)}
According to the \cite{schlichtkrull2018modeling}, the R-GCN update of an entity or node denoted by $v_i$ in a relational (directed and labeled) multi-graph is given as:
\begin{equation}
h_{i}^{(l+1)}=\sigma\left(\sum_{r \in \mathcal{R}} \sum_{j \in \tilde{\mathcal{N}}_{i}^{r}} \frac{1}{c_{i, r}} \mathbf{W}_{r}^{(l)} \mathbf{h}_{j}^{(l)}+\mathbf{W}_{0}^{(l)} \mathbf{h}_{i}^{(l)}\right),
\end{equation}
where $\tilde{\mathcal{N}}_{i}^{r}$ denotes the set of neighbor indices of node $i$ under relation $r \in R$, $c_{i,r}$ is a problem-specific normalization constant that can either be learned or chosen in advance \eg, $c_{i,r}=\left|\mathcal{N}_{i}^{r}\right|$.

Our Generalized Factorized Bilinear model with the summarization function, termed as GFB-R-GCN, becomes:
\begin{equation}
    \mathbf{h}^{(l)}_v = \sigma\Big(\sum_{r \in \mathcal{R}} \sum_{j \in \tilde{\mathcal{N}}_{i}^{r}}\mathbf{W}_{r}^{(l)} \mathbf{h}^{(l-1)}_j + \lambda_l\textbf{GenVec}(\mathbf{M})\Big),
\end{equation}
where we set $\mathbf{M}\!=\!\mathbf{W}_r^{(l)} \mathbf{h}^{(l-1)}_j  {\mathbf{h}^{(l-1)}_j}^\top\mathbf{W}_r^{(l)}\top$, and we choose a desired definition of $g(\cdot)$ according to the main submission \eg, we use the \textbf{MeanVec} operator.

% \begin{table}[]
% \centering
% \caption{Number of entities, relations, edges and classes
% along with the number of labeled entities for each of the
% datasets. Labeled denotes the subset of entities that have labels and that are to be classified.}
% \label{tab:kbdata}
% \resizebox{0.48\textwidth}{!}{ 
% \begin{tabular}{l|l|l|l|l}
% \hline
% Dataset   & AIFB   & MUTAG  & BGS     & AM        \\
% \hline
% Entities  & 8,285  & 23,644 & 333,845 & 1,666,764 \\
% Relations & 45     & 23     & 103     & 133       \\
% Edges     & 29,043 & 74,227 & 916,199 & 5,988,321 \\
% Labeled   & 176    & 340    & 146     & 1,000     \\
% Classes   & 4      & 2      & 2       & 11       
% \end{tabular}}
% \end{table}

\paragraph{Datasets.} We evaluate our model on four datasets in Resource Description Framework (RDF) format \cite{ristoski2016collection}: AIFB, MUTAG, BGS, and AM. Relations in these datasets need not necessarily encode directed subject-object relations, but are also used to encode the presence or absence of a specific feature for a given entity. For each dataset, the targets to be classified are properties of a group of entities represented as nodes.
%The exact statistics of the datasets can be found in Table~\ref{tab:kbdata}. 
For a more detailed description of the datasets the reader is referred to \cite{ristoski2016collection}. We remove relations that were used to create entity labels: employes and affiliation for AIFB, isMutagenic for MUTAG, hasLithogenesis for BGS, and objectCategory and material for AM.

\begin{table}[t]
\centering
\caption{Entity classification results (accuracy \% averaged over 10 runs) for a feature-based baseline (see the main text for details), WL  RDF2Vec , R-GCN and GFB-R-GCN . Test performance is reported on the train/test set splits provided by R-GCN.}
\label{tab:entityresult}
\begin{tabular}{|l|l|l|l|l|}
\hline
Model   & AIFB  & MUTAG & BGS   & AM    \\
\hline
Feat    & 55.55 & 77.94 & 72.41 & 66.66 \\
WL      & 80.55 & \textbf{80.88} & 86.20 & 87.37 \\
RDF2Vec & 88.88 & 67.20 & 87.24 & 88.33 \\
\hline
R-GCN   & 95.83 & 73.23 & 83.10 & 89.29 \\
GFB-R-GCN & \textbf{97.22} & 77.94 & \textbf{89.66} & \textbf{90.40} \\
\hline
\end{tabular}
\end{table}

\begin{table*}[]
  \centering
  \caption{Results for the problem of graph classification.}
    \begin{tabular}{rrrrrr}
    \toprule
          & MUTAG & PROTEINS & IMDB-Binary & Redit-Binary & COLLAB \\
    \midrule
    GCN   & 74.6 $\pm$ 7.7 & 73.1 $\pm$ 3.8 & 72.6 $\pm$ 4.5 & 89.3 $\pm$ 3.3 & 80.6 $\pm$ 2.1 \\
    GraphSage & 74.9 $\pm$ 8.7 & 73.8 $\pm$ 3.6 & 72.4 $\pm$ 3.6 & 89.1 $\pm$ 1.9 & 79.7 $\pm$ 1.7 \\
    \hline
    GFB-GCN   & 83.5 $\pm$ 4.7  & 74.2 $\pm$ 3.9 & 74.4 $\pm$ 5.4 & 90.6 $\pm$ 2.6 &  80.0 $\pm$ 2.3 \\
    GFB-GraphSage & 81.4 $\pm$ 7.5 & 74.4 $\pm$ 4.4 & 73.9 $\pm$ 4.9 & 88.9 $\pm$ 2.2 & 79.5 $\pm$ 2.3 \\
    \bottomrule
    \end{tabular}%
  \label{tab:GC}
\end{table*}

\paragraph{Baselines.}
As a baseline for our experiments, we compare against recent state-of-the-art classification results from RDF2Vec embeddings \cite{ristoski2016rdf2vec},
Weisfeiler-Lehman kernels (WL) \cite{shervashidze2011weisfeiler,de2015substructure}, and hand-designed feature extractors (Feat)~\cite{paulheim2012unsupervised}. Feat assembles a feature vector from the in- and out-degree (per relation) of every labeled entity. RDF2Vec extracts walks on labeled graphs which are then processed using the Skipgram~\cite{mikolov2013distributed} model to generate entity embeddings,
used for subsequent classification. All entity classification experiments were performed on a CPU machine with 64GB of memory.

\paragraph{Results.}
Results in Table~\ref{tab:entityresult} are reported on the train/test benchmark splits from \cite{ristoski2016collection}. We further set aside 20\% of the training set as a validation set for hyperparameter tuning. For R-GCN, we report performance of a 2-layer model with 16 hidden units (10 for AM). We trained with Adam \cite{kingma2014adam} for 50 epochs using a learning rate of 0.01. The normalization constant is chosen as $c_{i, r}=\left|\mathcal{N}_{i}^{r}\right|$. %Further details on (baseline) models
and hyperparameter choices are provided in the supplementary material. 
The GFB-GCN employ the \textbf{MeanVec} in the aggregator function.

\section{Graph Classification}
In Table \ref{tab:GC}, we report the average accuracy of 10-fold cross validation on a number of common benchmark datasets. We randomly sample a training fold to serve as a validation set. We only make use of discrete node features. In case they are not given, we use one-hot encodings of node degrees as feature input. For all experiments, we use the global mean operator to obtain graph-level outputs.
For evaluating the Generalized Bilinear Pooling, we use the GraphSAGE and GCN as our baseline. For each dataset, we tune (1) the number of hidden units
$\in \{16, 32, 64, 128\}$ and (2) the number of layers $\in \{1, 2, 3, 4, 5\}$ with respect to the validation set. Our GFB based methods use \textbf{DiagVec} in the aggregation function.

\bibliography{gfba}
\bibliographystyle{plain}

\end{document}